\documentclass[review]{elsarticle}

\usepackage{lineno,hyperref}
\modulolinenumbers[5]

\journal{Journal of \LaTeX\ Templates}

\usepackage{amsmath}
\usepackage{booktabs}
\usepackage{float}
\usepackage{placeins}

\usepackage{xcolor}

\usepackage[compact]{titlesec}         % you need this package
\titlespacing{\section}{2pt}{2pt}{2pt} % this reduces space between (sub)sections to 0pt, for example

\begin{document}

\begin{frontmatter}

\title{Transformer-Based Approach for Joint Handwriting and Named Entity Recognition in Historical documents}

%% \author{Ahmed Cheikh Rouhou}
%% \address{Radarweg 29, Amsterdam}
%% \fntext[myfootnote]{Since 1880.}

\author[firstaddress]{Ahmed Cheikh Rouhou\corref{mycorrespondingauthor}}
\cortext[mycorrespondingauthor]{Corresponding author}
\ead{a.cheikhrouhou@instadeep.com}

\author[firstaddress,secondyaddress,thirdaddress]{Marwa Dhiaf}

\author[secondyaddress,thirdaddress]{Yousri Kessentini}
\author[firstaddress]{Sinda Ben Salem}

\address[firstaddress]{InstaDeep}
\address[secondyaddress]{Digital Research Center of Sfax}
\address[thirdaddress]{SM@RTS : Laboratory of Signals, systeMs, aRtificial Intelligence and neTworkS}

\begin{abstract}
The extraction of relevant information carried out by named entities in handwriting documents is still a challenging task. Unlike traditional information extraction approaches that usually face text transcription and named entity recognition as separate subsequent tasks, we propose in this paper an end-to-end transformer-based approach to jointly perform these two tasks. The proposed approach operates at the paragraph level, which brings two main benefits. First, it allows the model to avoid unrecoverable early errors due to line segmentation. Second, it allows the model to exploit larger bi-dimensional context information to identify the semantic categories,  reaching a higher final prediction accuracy. We also explore different training scenarios to show their effect on the performance and we demonstrate that a two-stage learning strategy can make the model reach a higher final prediction accuracy.
As far as we know, this work presents the first approach that adopts the transformer networks for named entity recognition in handwritten documents. We achieve the new state-of-the-art performance in the ICDAR 2017 Information Extraction competition using the Esposalles database,  for the complete task,  even though the proposed technique does not use any dictionaries, language modeling, or post-processing. 
\end{abstract}

\begin{keyword}
Named-Entity-Recognition \sep Text Block Recognition \sep Transformer
\sep IEHHR Competition
\end{keyword}

\end{frontmatter}

%\linenumbers
\section{Introduction}
In the last decades, researchers have been exploring various document recognition techniques to recover textual information from images. Lately, optical character recognition techniques have achieved high accuracy in recovering texts from modern documents. However, they need some refinement while handling historical documents due to the degraded quality of the images and the complexity of the old handwriting styles. 

Although Handwritten Text Recognition (HTR) of historical document images is a good step to recover textual information \cite{9413255}, there is an increasing interest within the research community regarding information extraction and document understanding to allow meaningful semantic access to the information contained in document collections.

In this context, Named Entity Recognition (NER) from document images is one of the most challenging and practical problems, which consists of transcribing textual contents and classifying them into semantic categories (names, organizations, locations, etc). 

In the literature, traditional NER methods on document images mainly adopt two processing steps \cite{8791217,Ruokolainen2020NameTN,dinarelli-rosset-2012-tree-structured,10.1145/3352631.3352637}. Text information is extracted firstly via the HTR process, and then Natural Language Processing (NLP) techniques are applied to parse the output text and extract the named entity tags. Despite the recent improvements of deep learning-based NLP systems, the performance of these two-stages approaches still relies on the quality of the HTR processing step. Generally, errors of the HTR stage due to the low-quality scans, for example, affect the NLP stage’s performance considerably.

The second category aims to jointly perform transcription and named entity recognition from the document images without an intermediate HTR stage \cite{8395229,CARBONELL2020219,Toledo2016HandwrittenWI,TOLEDO201927}. Most studies of this second category confirm the benefit when leveraging the dependency of these pairs of tasks with a single joint model. In \cite{Toledo2016HandwrittenWI} a single Convolutional Neural Network (CNN) is used to directly classify word images into different categories skipping the recognition step. However, this approach does not use the context surrounding the word to be classified, which might be critical to correctly predict named entity tags. In \cite{TOLEDO201927} a CNN is combined with a Long Short-Term Memory (LSTM) network to integrate a larger context, achieving better results compared to \cite{Toledo2016HandwrittenWI}. Still, in this work, the context is limited to the line level, which affects the extraction of semantic named entity tags. To integrate a bi-dimensional context, authors in \cite{CARBONELL2020219} propose an end-to-end model that jointly performs handwritten text detection, transcription, and named entity recognition at the page level, capable of benefiting from shared features for these tasks. This approach presents two main drawbacks. First, it requires word bounding box annotation, which is a huge cost saving in the real application. Second, the proposed multi-task model can be limited in performance in cases where one specific task is much harder and unrelated to the others.

Recently, inspired by their success in many NLP applications, Sequence-to-Sequence (Seq2Seq) approaches using attention-based encoder-decoder architectures have started to be successfully applied for HTR \cite{10.5555/3305381.3305483,8978104}. Most of these architectures still combine the attention mechanism with a recurrent network (BLSTMs or GRU) which severely affects the effectiveness when processing longer sequence lengths by imposing substantial memory limitations. Recently, authors in \cite{kang2020pay} propose an architecture inspired by transformers, which dispenses any recurrent network for HTR of text-line images. The major drawback of this method is that the line segmentation errors are often irreversible and will therefore significantly affect the recognition performance.

For handwritten historical document recognition, the line segmentation process is a complicated task compared to modern documents. Besides the complexity of handwritten texts (inconsistent spaces between lines, characters of successive lines can be overlaid, etc.), text images can involve distortions and noisy pixels due to the quality of these documents. Many studies tried to enhance the quality of the document image before the segmentation \cite{9187695} or to improve the segmentation quality in historical documents \cite{8978147,9257752,XIE2019271}. However, in most cases, the segmentation is applied as a preprocessing step before recognition.   
%10.5555/3157096.3157190
Lately, researchers have been examining the recognition of text blocks instead of text lines without any segmentation step \cite{8270105,coquenet2020endtoend}, following two categories of approaches. In the first category, the text-block images are transformed into lines representation using convolution layers \cite{yousef2020origaminet} or attention mechanism \cite{coquenet2020endtoend}, in order to perform Connectionist Temporal Classification (CTC) decoding. In the second approach, feature extraction conserves the 2D representation of the text block, then, the decoding is performed using 2D-CTC \cite{8395230} or attention-based Seq2Seq architecture \cite{8270105}. As far as we know, there have been no works in the literature applying the transformer architecture at paragraph level  to perform jointly HTR and NER. 

Motivated by the above observations, we propose in this paper an end-to-end transformer-based approach to jointly perform full paragraph handwriting and named entities recognition in historical documents. To the best of our knowledge, this is the first study that involves the transformer architecture  \cite{10.5555/3295222.3295349} for such a task. The aim is to surpass the line segmentation problems, as well as to allow the model to exploit larger bi-dimensional context information to identify the semantic NE tags. To this end, our first contribution consists in adapting the transformer architecture to deal with the 2D representation of the input text block. For this aim, the 2D features maps obtained by the ResNet architecture are transformed into 1D  sequential  features  using  flattening operation. In order to add positional information, we have tested two positional encoding (PE) methods: the 2D-based PE which is performed on the 2D features maps and the 1D PE applied to the 1D feature sequence. 

The second contribution of this paper consists in exploring different training scenarios including two-stage learning, mixed-data learning, and curriculum learning to show their effect on the performance and make the model reach a higher final prediction accuracy. %Extensive ablation and comparative experiments are conducted  to  validate  the  effectiveness  of  our  approach. \\

The major contributions of this paper can therefore be summarized as follows:
\begin{itemize}
    \item We propose an end-to-end transformer-based architecture for named entity recognition in handwritten document images.
    \item The proposed method jointly performs handwriting and named entity recognition at the paragraph level, allowing the model to avoid unrecoverable early errors due to line segmentation and to exploit larger context information to identify the semantic relations between the named entities.
    \item We explore different training scenarios including two-stage learning, mixed-data learning, and curriculum learning to show their effect on the performance and make the model reach a higher final prediction accuracy. 
    \item Extensive ablation and comparative experiments are conducted to validate the effectiveness of our approach. Even though the proposed technique does not use any dictionaries, language modeling, or post-processing, we achieve new state-of-the-art performance on the public IEHHR competition \cite{8270158}.
\end{itemize}

The remaining part of this paper is structured as follows. The approach is presented in section 2 including a description of the proposed architecture and relatively proposed methods. Section 3 displays the experimental results. The conclusion and perspectives are stated in section 4. 

\section{Proposed Approach}
\begin{figure*}
  \includegraphics[width=0.95\textwidth]{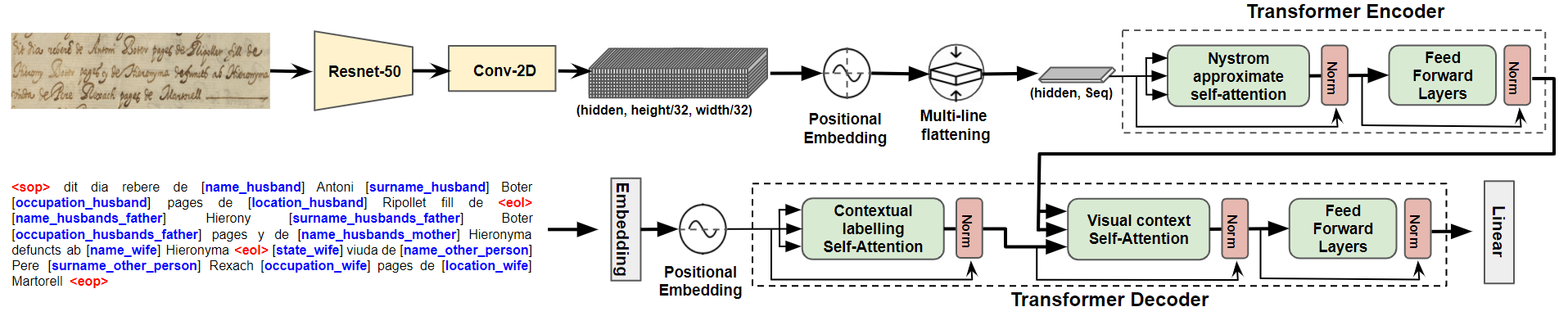}
  \caption{Overview of the proposed architecture.}\label{fig1}
\end{figure*}

The proposed approach consists of creating an end-to-end neural network architecture, shown in figure \ref{fig1}, that recognizes texts and possible named entities from multi-line historical images. This network is composed of two components: multi-line feature extraction and transformer-based sequence labeling. 

The input image is fed to ResNet-50 architecture to extract features. The number of obtained features is compressed to match the transformer parameters using a simple 2D convolutional layer. After, the 2D positions encoding is merged with these features to add positional information to each feature vector. Next, the 2D features map is transformed into 1D sequential features using flatten function. At this level, this sequence matches the requirement of the transformer to perform sequence labeling. The transformer encoder performs a multi-head self-attention mechanism on the visual-features-sequence. Finally, the ground truth is introduced to an embedding layer with a 1D positional encoding technique then to the transformer decoder, where visual encoded features are matched with the ground truth. The output sequence of the architecture is obtained via a linear layer transforming the transformer hidden size to the number of labels.

\subsection{Multi-line feature extraction}
To perform HTR and named entity recognition, text images need to be transformed into a high-level feature representation. In this approach, we deal with paragraph-level images, where the number of lines per paragraph is unknown. 

The input image is converted to a gray-scale level as color information does not have an influence on the textual entities in images.  Moreover, we duplicate the image to match the three-channel dependency needed by the architecture. Next, the image is resized to match the proposed architecture with keeping aspect-ratio. The resize operation is not mandatory, but experiments have proved that image size has a huge influence on the results. Afterward, we have chosen to use ResNet-50 \cite{7780459} as our feature extractor, also called backbone architecture. The ResNet produces a compact feature representation ($H'=\frac{height}{32}$ and $W'=\frac{width}{32}$) with a contextualized view of the full image. 
%At this stage, the resulting 3D features representation shape can be described as detailed in equation \ref{eq1}.
%\vspace{-2mm}
%\begin{equation}
%\label{eq1}
%F_{ResNet}=(2048,\frac{height}{32},\frac{width}{32})
%\end{equation}
The lines' information is conserved by their corresponding receptive fields (a correspondence between a feature vector and the input image) in this 3D features representation.   

%\begin{figure}
%  \includegraphics[width=0.46\textwidth]{Fig_02.png}
%  \caption{Receptive fields representation.}\label{fig2}
%\end{figure}

In order to make the features size correlated with the hidden-size (hidden) hyper-parameter of the transformer, we added a 2D-convolutional layer ($Conv2D$), preserving the same height/width dimensions of the input, that transformers the 2048-based features to match the hidden-size of the transformer. 
%\vspace{-2mm}
%\begin{equation}
%\label{eq2}
%F_{Conv2D}=(h,\frac{height}{32},\frac{width}{32})
%\end{equation}

To bypass the need for any recurrence in the network architecture, and Adaptive 2D Position Encoding (A2DPE) \cite{9151002} is applied, to add 2D positional information to the features without modifying their shape. The A2DPE adapts the original position encoding algorithm in \cite{10.5555/3295222.3295349} to encode two-dimensional inputs, which fits best with our proposed approach.  The positional encoding is applied as
%For the sake of clarity, the receptive fields of the features represent characters in different text lines, thus, using 2D position information enhances the recognition process of each character-related feature with its corresponding line.

%\vspace{-1mm}
%\begin{equation}
%\label{eq3}
%F_{A2DPE}=A2DPE(F_{Conv2D})
%\end{equation}

\begin{equation}
\label{eq3}
\begin{split}
& A2DPE(E)=\{E_{h,w}+POS_{h,w}\} \\
& \mbox{for } h\in[1..H'] \mbox{ and } w\in[1..W']
\end{split}
\end{equation}

where $E_{h,w}$ represents a feature vector at position $(h,w)$ obtained from $Conv2D$ and $POS_{h,w}$ are the positional encoding of that features. The latter is computed using two scale factors $\alpha(E)$ and $\beta(E)$ obtained by applying global average pooling $g(E)$ followed by 2-layer perceptron per dimension $(W_1^h,W_2^h,W_1^w,W_2^w)$: 

\begin{equation}
\label{eq3.1}
\begin{split}
& \alpha(E)=sigmoid(max(0,g(E)W_1^h)W_2^h), \\
& \beta(E)=sigmoid(max(0,g(E)W_1^w)W_2^w)
\end{split}
\end{equation}
and the positional encoding is computed as

\begin{equation}
\label{eq3.2}
POS_{h,w}=\alpha(E)\widetilde P_h+\beta(E)\widetilde P_w
\end{equation}
$\widetilde P$ is a sinusoidal encoding over height ($\widetilde P_h$) and width ($\widetilde P_w$) as described in Eq. \ref{eq3.3}:
\begin{equation}
\label{eq3.3}
\begin{split}
& \widetilde P_{p,2i} = sin(p/10000^{2i/D}), \\
& \widetilde P_{p,2i+1} = cos(p/10000^{2i/D})
\end{split}
\end{equation}
where $p$ and $i$ are indices along position and hidden dimensions, respectively. \\

The final step is to transform the 2D representation of the features into a simple features sequence as the transformer requires. This can be accomplished by flattening the height and width dimensions of the 3D features representation resulting into a 2D representation of the features.%, as described in equation \ref{eq4}.\\

%\vspace{-3mm}
%\begin{equation}
%\label{eq4}
%\begin{split}
%& F_{Flatten}=Flatten(F_{A2DPE})=(h,Seq_{len})\\
%& \mbox{where } Seq_{len}=\frac{height}{32} \times %\frac{width}{32}
%\end{split}
%\end{equation}

After performing these steps into a paragraph-level image, we obtain a features sequence of the whole image including character-wise and line-wise positional information per frame. 
% Table \ref{tab0} shows output shapes and parameters of different layers leading to the sequential representation of the features.

%\begin{table}[]
%\caption{Multi-line feature extraction architecture details.}
%\centering
%\begin{tabular}{ccc}
%\hline
%\textbf{Layer} & \textbf{Output Shape} & \textbf{Parameters} \\ \hline
%$Input$          & (3,$h$,$w$)               & -                   \\
%$ResNet-50$      & (2048,$h'$,$w'$)          & 23.6M               \\
%$Conv2D$         & (hidden,$h'$,$w'$)        & 1.0M                \\
%$A2DPE$          & (hidden,$h'$,$w'$)        & 1.1M                \\
%$Flatten$        & (hidden,$h'\times w'$)        & -                   \\ \hline
%\end{tabular}
%\label{tab0}
%\end{table}
%

\subsection{Transformer-based sequence labeling}
In our approach, we utilize the transformer architecture proposed in \cite{10.5555/3295222.3295349} to transform images into transcription containing characters and named entities. The transformer has two modules: a multi-head self-attention-based feature-encoder and a multi-head self-attention decoder that transcribes the characters and named entities using the encoded features. 

\subsubsection{Feature sequence encoding}
The main objective of this encoder is to prepare a more effective representation of the features according to the label transcriber. The multi-head self-attention is applied to the flattened feature sequence to add more context and relations between the sequence’s frames. 

%\begin{figure}
%\centering
%  \includegraphics[width=0.41\textwidth]{Fig_03.png}
%  \caption{ Representation of different feature vectors of a single character.}\label{fig3}
%\end{figure}

This encoding step coordinates the different features vectors that represent a single character or a named entity, in the input sequence. The self-attention mechanism assures this correlation within the features as described in equation \ref{eq5}.
\begin{equation}
\begin{split}
\label{eq5}
F_{Encoder}=\{v^0,v^1,..v^{Seq_{Len}-1}\}\\
\mbox{and } v^i=Softmax(\frac{q_i \times K}{hidden}) \times V
\end{split}
\end{equation}

where $Seq_{len}=h' \times w'$,  $K$ represents the key, $V$ represents the value, $q_i \in Q$ and $i=\{0,..,Seq_{len}-1\}$ represents the query. These representations are a learned parameters and have the same shape of the input feature sequence.
\subsubsection{Text and named entity decoder}
The text and named entities are recognized using the second part of the transformer architecture denoted as the decoder. This decoder learns the correlation between visual encoded features and their corresponding labels.

There are two types of labels in our approach: visual labels and contextual labels. The visual labels ($V_{labels}$) are the readable characters from texts in images. The contextual labels ($C_{labels}$) are the start of paragraph ($<sop>$), the end of line ($<eol>$), the end of paragraph $<eop>$, and named entities tags ($[tag]$). All these labels represent the number of classes ($NB_{class}$) of the output layer as equation \ref{eq7}.

\begin{equation}
\label{eq7}
NB_{class}=\{ V_{labels} \} \cup \{C_{labels} \}
\end{equation}

\begin{figure}[]
\centering
  \includegraphics[width=0.41\textwidth]{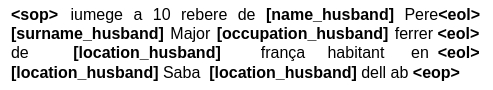}
  \caption{Example of the textual input of the Transformer decoder.}\label{figIcon}
\end{figure}

The input of the decoder,  as shown in figure \ref{figIcon},  is the textual contents of the image surrounded by $<sop>$ and $<eop>$ tags, and contains the possible tags (placed before their relative word). After preparing the input text, the characters are embedded to match the hidden-size dimension. Then, the final input of the transformer decoder ($F_{labels}$) is prepared by applying 1D Positional encoding (1DPE) \cite{10.5555/3295222.3295349} to the embedded labels.
%The shape of $F_{labels}$ is $(hidden,N)$ where $N$ represents the length of the input labels sequence.
The transformer architecture allows concurrent access to the sequence information, due to the self-attention mechanism involved, which facilitates the multi-line decoding of the visual encoded features. In the recurrent networks, the decoding steps follow the order of the visual features while including forward and backward contextual information. 

The transformer decoder involves two multi-head self-attention modules: the label self-attention module which learns the relation between different characters and tags, and the recognition self-attention module where labels are aligned to their corresponding visual encoded features. Finally, a linear layer transforms the output of the decoder into classes ($NB_{class}$), followed by $Softmax$ and $Arg-Max$ operations to select the right labels.

\subsection{Named entities encoding}
In the previous section, we have shown that labels of the system include named entity tags in the $C_{labels}$ list. These tags are inserted after some words corresponding to the named entities so that the model learns to predict them. Given a paragraph composed of $N$ words, $M$ tags, and $L$ lines, the encoding is presented as shown in figure \ref{figIcon}.

%\vspace{-1mm}

In this study, two types of named entities are available: category-related and person-related tags. The category tags ($C_{tags}$) denote the semantic of the named entity (name, occupation, location, etc). The person tags ($P_{tags}$) provide the person associated to each category (husband, wife, wife's father, etc). The full list of tags is described in \cite{8270158}.

we explore in this work two encoding strategies for the NE tags like in \cite{8395229}:
\begin{enumerate}
    \item Single Separate tagging: the category and person tags are embedded as a single tag per type i.e. [name] [husband]. As a result, each named entity should have two tags. The total number of tags is :
    \[ NB_{class} = |C_{tags}| \cup  |P_{tags}| \cup  V_{labels}  \]
    \item Joint tagging: a single tag is assigned to each named entity, where the tag contains the category and the person information i.e. [name\_husband]. The total number of tags is :
    \[ NB_{class} = (|C_{tags}| \times  |P_{tags}|) \cup V_{labels}   \]
\end{enumerate}
Following this encoding, the model learns to predict NE tags without any post-processing.
\subsection{Two-stage learning}
In this work, we propose to train the architecture using a two-stage learning approach: the first stage consists of training the model for HTR only, i.e. named entity tags are not included in the ground truth. We call this stage pre-training, where the model learns how to recognize the character labels and $<eol>$ labels without any knowledge of the named entity tags.

In the second stage, the model is finetuned using the full representation of the ground truth including named entity tags. We refer to this stage as fine-tuning. After applying this stage, the model can perform the recognition of characters and named entity tags.
\subsection{Mixed-level learning vs Curriculum learning}
The next variable we examined was the level of input images in the training. The simplest strategy consists of training the proposed transformer-based model directly at the paragraph level. Nevertheless, we believe that training it on different levels of input (1-line bloc, 2-lines, ..., paragraph) can help the model to learn the reading order independently of the number of lines in the bloc. For this aim, we propose two different scenarios named mixed-level learning and curriculum learning. In the mixed-level learning scenario, the model is learned in one stage on a collection of blocs containing different numbers of lines. In the curriculum learning scenario, our model starts with learning easy examples (line text images) then progressively increases the number of lines in the bloc (2-lines, 3-lines, etc) until we reach the paragraph-level images. Humans have applied this concept for decades in any learning strategy. Two curriculum-learning strategies are proposed in this work. The first one, denoted curriculum sequential learning, consists of training the model progressively on text transcription (without NE tags) using different levels of input (1-line bloc, 2-lines, ..., paragraph). When the training reaches the paragraph-level images, a final fine-tuning is performed to learn the named entity tags at the paragraph-level image. The second strategy denoted curriculum dual learning, where the model alternates the training on text transcription and NE tags at each increased level of input. 

\section{Experiments}
\subsection{Datasets}
\textbf{Esposalles dataset:}
We have conducted our experiments on the public dataset proposed in ICDAR 2017 Information Extraction from Historical Handwritten Records (IEHHR) competition \cite{8270158}. This dataset is a subset of the Esposalles dataset \cite{ROMERO20131658}. It has been labeled for information extraction. The dataset collects 125 handwritten pages, containing 1221 marriage records (paragraphs). Each record is composed of several text lines giving information of the husband, wife, and their parent’s names, occupations, locations, and civil states.  We have used  872 records for training, 96 records for validation, and 253 records for evaluation.\\

\begin{figure}[h]
\centering
  \includegraphics[width=0.4\textwidth]{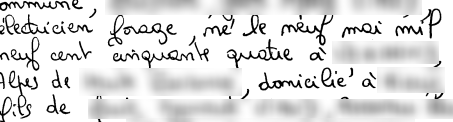}
  \caption{ Sample image from the FHMR dataset. Some words are blurred for privacy.}\label{figstudia}
\end{figure}

\textbf{FHMR dataset:}
The second dataset used in this study is French Handwritten  Marriage Records (FHMR) \cite{FHMR}. FHMR is a private dataset that provides a collection of  French handwritten marriage records. Each record contains several text lines giving information of the wife, husband, and their parent’s names, occupations, locations, and dates. The text images are provided at the line and paragraph levels. An example of a paragraph image is shown in Figure \ref{figstudia}.  We have used  997, 103, 132 records for the training, the validation, and the evaluation, respectively.

% \begin{table}[!h]
%\centering
%\caption{Overview of the FHMR dataset.}
%%\begin{tabular}{cccc}
%\hline
%\textbf{Image-level} & \textbf{Train} & \textbf{Validation} & \textbf{Test} \\ \hline
%Pages                & 88             & 11                  & 13            \\
%Records              & 997            & 103                 & 132           \\
%Lines                & 2376           & 286                 & 546           \\ \hline
%\end{tabular}
%\label{tab1}
%\end{table}
%\vspace{-12pt}
\subsection{Metrics}
In this study, we utilize the same metrics used in the IEHHR competition. This allows us to fairly compare our system with the proposed ones in the competition. The main objective of the metric used is to prioritize the category tags, at a first level, the person tags, at a second level, then text recognition accuracy at a final level. We have used the basic and complete score in our study: the basic score is calculated using the category tag and text recognition accuracy, and the complete score considers all the tags and text recognition accuracy. More details about the metrics can be found in \cite{8270158}.

\subsection{Ablation study}
To choose the best hyper-parameters of the proposed model, we have performed an ablation study online and paragraph images using the joint tagging encoding of the ground truth. The study includes a variation of the input image size, positional encoding algorithm, hidden size, heads, and layers of the transformer.  For the positional encoding operation, we have tested also the 1DPE algorithm to encode the feature map. This algorithm is applied after the multi-line flattening operation, where features are represented as a 1D sequence. 

\begin{table*}[!htb]
\centering
\caption{Ablation study on different hyper parameters of the proposed architecture.}
\begin{tabular}{cccccccc}
\hline
\textbf{Image Level} & \textbf{Image Size} & \textbf{Positional Encoding} & \textbf{Hidden Size} & \textbf{Heads} & \textbf{Layers} & \textbf{Basic} & \textbf{Complete} \\ \hline
Paragraph*           & 128x512             & 1D                           & 512                  & 1              & 1               & 56.31          & 48.93             \\
Paragraph*           & 256x1024            & 1D                           & 512                  & 1              & 1               & 93.24          & 92.31             \\
%Paragraph            & 256x1024            & 1D                           & 512                  & 1              & 1               & 94.51          & 93.77             \\
%Paragraph            & 256x1024            & 2D                           & 512                  & 1              & 1               & 94.61          & 93.95             \\
Paragraph            & 256x1024            & 2D                           & 512                  & 1              & 2               & 92.65          & 92.41             \\
Paragraph            & 256x1024            & 1D                           & 256                  & 1              & 2               & 95.72          & 95.72             \\
\textbf{Paragraph}   & \textbf{256x1024}   & \textbf{2D}                  & \textbf{256}         & \textbf{1}     & \textbf{2}      & \textbf{95.74} & \textbf{95.12}    \\
%Paragraph            & 256x1024            & 2D                           & 256                  & 1              & 3               & 86.37          & 84.74             \\
Paragraph            & 256x1024            & 2D                           & 256                  & 2              & 2               & 32.85          & 26.94             \\
Paragraph            & 256x1024            & 2D                           & 256                  & 4              & 4               & 20.04          & 12.35             \\
Paragraph            & 384x1024            & 1D                           & 256                  & 1              & 2               & 55.44          & 47.81             \\\hline
%Line                 & 32x512              & 1D                           & 256                  & 1              & 2               & 87.99          & 86.42             \\
Line                 & 128x1024            & 1D                           & 256                  & 1              & 2               & 94.12          & 91.78             \\
%Line                 & 256x1024            & 1D                           & 256                  & 1              & 2               & 94.54          & 91.54             \\
\textbf{Line}        & \textbf{256x1024}   & \textbf{2D}                  & \textbf{256}         & \textbf{1}     & \textbf{2}      & \textbf{95.16} & \textbf{93.3}     \\ \hline
\end{tabular}
\label{tab2}
\end{table*}
%\vspace{-2mm}

Table \ref{tab2} shows the most important results of the ablation study. We have eliminated most of the experiments made with image size greater than 384$\times$1024, heads $>1$, layers $>2$ as they all led to lower recognition rates. The symbol (*) in the table indicates that the used dataset does not integrate the label end-of-line $<eol>$ in the ground truth transcription.

The best performance is achieved using an architecture taking as input an image of size 256$\times$1024 at both paragraph and line levels as shown in the table \ref{tab2}. We observe also that 2D positional encoding performs slightly better than 1D positional encoding at both paragraph and line levels.  Concerning the hyper-parameters, the best performance is obtained using 256 hidden sizes i.e. feature vector and self-attention sizes, 1 attention-head, and 2 layers. We also observe that the model performs better when the image input is a paragraph for both tasks (basic, complete) which confirms that it is more difficult to predict semantic NE while considering only a single line image, due to lack of context. Note that integrating the $<eol>$ label in the ground-truth transcription improves the performance of the model. 

For the next experiments, we keep the best hyper-parameters configuration to investigate the different ways of encoding the image transcripts and semantic labels and the different learning scenarios.  
\subsection{Results of different NE encoding}
As explained in section 2, our study is conducted using the different encoding of the image transcripts and semantic labels including single tags and joint tags representation.

As presented in table \ref{tab3}, the joint tagging gave better results than the single tagging despite the highest number of tags. As an explanation, it is simpler for the decoder to choose a combined tag instead of two separated tags. Also, using a single tag approach, the system wrongly detects more than two tags per named entity.

%\vspace{-12pt}
\begin{table}[htp]
\centering
\caption{Paragraph-level results on different NE.}
\begin{tabular}{ccc}
\hline
\textbf{Learning method}     & \textbf{Basic} & \textbf{Complete} \\ \hline
Joint tags representation    & 95.74          & 95.127            \\
Separate tags representation & 93.2           & 91.5              \\ \hline
\end{tabular}
\label{tab3}
\end{table}
%\vspace{-12pt}
\subsection{Results of the different learning strategies}
We also investigate multiple training strategies including one-stage learning, two-stage learning, mixed-level learning, curriculum sequential learning and curriculum dual learning. 

%\vspace{-12pt}
\begin{table}[htp]
\centering
\caption{Results of the different learning strategies.}
\begin{tabular}{ccc}
\hline
\textbf{Learning method}       & \textbf{Basic} & \textbf{Complete} \\ \hline
 One-stage                      & 95.74          & 95.127             \\
Two-stage                      & 96.02          & 95.39             \\
Mixed-level                    & 95.67          & 95.52             \\
\textbf{Two-stage mixed-level} & \textbf{96.25} & \textbf{95.54}    \\
Curriculum sequential          & 95.5           & 94.86             \\
Curriculum dual                & 95.33          & 94.8              \\ \hline
\end{tabular}
\label{tab4}
\end{table}

%\vspace{-12pt}
As shown in Table \ref{tab4}, the two-stage learning (learning on text transcription then fine-tuning on NE tags) improves the performance compared to the one-stage learning (learning directly on text transcription and  NE tags). An improvement of $0.28\%$ on the basic score and of $0.263\%$ on the complete score is obtained.

The next variable we investigated is the level of input images in the training. As shown in table \ref{tab4}, the best performance is achieved with mixed-level training with a basic score of $96.25\%$ and a complete score of $95.54\%$. We notice also that both scenarios of curriculum learning (dual and sequential) do not improve the performance compared to the mixed-level training. We believe that these two scenarios can be more useful when dealing with larger training datasets.  

\subsection{Comparaison with IEHHR Competition}
We compared our proposed approach with other methods from the literature that participated in the IEHHR competition \cite{8270158}. The comparison is straightforward as we use the same experimental protocol used in the competition to evaluate our system. Table \ref{tab5} shows the performance results of some methods from the literature and our best results on line-level and paragraph level. 
 As reported in table \ref{tab5}, our model presents the best performance with a gain of respectively $+2.09\%$ and $+3.57\%$ on basic and complete scores, compared to the best method  performing the complete tag recognition task in the competition \footnote{https://rrc.cvc.uab.es/?ch=10\&com=evaluation\&task=1}. This result confirms the suitability of the proposed approach even though it does not use any dictionaries, language modeling, or post-processing.

%\vspace{-12pt}
\begin{table}[!h]
\centering
\caption{Comparison with  IEHHR Competition systems.}
\begin{tabular}{cccc}
\hline
\textbf{System} & \textbf{Basic} & \textbf{Complete} & \multicolumn{1}{l}{\textbf{Level}} \\ \hline
Hitsz-ICRC-2$^*$ &  94.16       &   91.97          &     Word \\
Baseline-CNN$^*$ &  79.40       &   70.18          &     Word  \\
CITLab-Argus-1$^*$      &  89.53       &   63.08          &     Line                       \\
CITLab-Argus-2$^*$      &  91.93       &   91.56          &     Line                       \\
CITLab-Argus-3$^*$      &  91.61       &   91.17          &     Line                       \\
\cite{8395229}          &   90.58             &     89.39              &   Line                         \\
HMM-MGGI$^*$  &  80.28     &    63.11      &     Line                                  \\
Ours           &  95.16       &    93.3            &    Line                          \\
\textbf{Ours}             & \textbf{96.25}           & \textbf{95.54}              & Record                               \\ \hline
\end{tabular}
\label{tab5}\\
\footnotesize{$^*$ System mentioned in \cite{8270158}}
\end{table}

%\vspace{-12pt}
\subsection{FHMR Results}
The interest of this second experiment is to confirm the efficiency of our method when dealing with more complex document images with different languages using the challenging FHMR dataset. We compare our method to three different approaches: the first one is based on the CRNN model (similar to \cite{8395229}) that jointly performs transcription and semantic annotation of handwritten text images. The second approach relies on two stages, the first stage employs CRNN to transcribe documents into an electronic text, while the second seeks to locate semantic entities in text using the French version of BERT named CamemBERT \cite{cBERT}. The last one is also based on two stage approach and uses Transformer architecture for text transcription instead of CRNN. 

%\vspace{-12pt}
\begin{table}[htp]
\centering
\caption{FHMR results.}
\begin{tabular}{@{}llll@{}}
\toprule
Learning method         & Basic & Complete & Level  \\ \midrule
Ours: Transformer joint                    & 86.11 & 77.26    & Record \\
Transformer + CamemBERT & 76.52 & 72.50    & Record \\
CRNN joint              & 74.98 & 73.22    & Line   \\
CRNN + CamemBERT        & 72.18 & 71.00    & Line   \\ \bottomrule
\end{tabular}
 
\label{tab6}
\end{table}
%\vspace{-12pt}

Compared to the "CRNN joint" model, "Transformer + Camembert" model achieves a better performance for the basic task, which is the simplest one. This can be explained by a better transcription quality of the transformer compared to CRNN. On the contrary, for the complete task, which is more complex as the semantic category of NE need to be identified, "CRNN joint" model achieves a slightly better performance  confirming the interest of the joint training  to identify  the  semantic  relations between the named entities. As shown in table \ref{tab6}, the best performance is obtained by our  proposed "joint  Transformer"  model as it benefits from the high recognition quality of Transformer and the joint two-stage mixed-level training. 

\section{Conclusion}
In this paper, we have proposed an end-to-end architecture to perform paragraph-level handwriting and named entity recognition on historical document images. As far as we know, it is the first approach that adopts the transformer networks at paragraph level for such a task. In contrast to traditional approaches which are based on two subsequent tasks (HTR and NLP), the proposed method jointly learns these two tasks on only one stage. Detailed analysis and evaluation are performed on each module, confirming the suitability of the proposed method by capturing more contextual information. Indeed, the presented results prove that our method achieves state-of-the-art performance even though it does not include any dictionaries, language modeling, or post-processing. 
In our future work, we can scale the proposed approach to work on page-level images, where recent alternatives of the transformer have been proposed (TransformerXL, Performer, etc). Also, we can investigate the impact of using the transformer encoder directly on the image RAW representation without including any backbone architecture.

%\section*{References}

\bibliography{mybibfile}

\end{document}